\documentclass[nouppercase]{kaia}
\usepackage{amsmath}
\usepackage{booktabs}
\usepackage{xcolor}
\usepackage{amssymb}
\usepackage{makecell}
\title{Selective Aggregation of Attention Maps Improves Diffusion-Based Visual Interpretation}

\author{Jungwon Park}{a}
\author{Jungmin Ko}{b}
\author{Dongnam Byun}{a}
\author{Wonjong Rhee}{b}
\affiliation{Department of Intelligence and Information, Seoul National University, Seoul, Republic of Korea, \{quoded97, east928\}@ snu.ac.kr}{a}
\affiliation{Interdisciplinary Program in Artificial Intelligence, Seoul National University, Seoul, Republic of Korea, \{jungminko, wrhee\}@snu.ac.kr}{b}

\begin{document}

\maketitle
\begin{abstract}
	Numerous studies on text-to-image (T2I) generative models have utilized cross-attention maps to boost application performance and interpret model behavior. However, the distinct characteristics of attention maps from different attention heads remain relatively underexplored. In this study, we show that selectively aggregating cross-attention maps from heads most relevant to a target concept can improve visual interpretability. Compared to the diffusion-based segmentation method DAAM, our approach achieves higher mean IoU scores. We also find that the most relevant heads capture concept-specific features more accurately than the least relevant ones, and that selective aggregation helps diagnose prompt misinterpretations. These findings suggest that attention head selection offers a promising direction for improving the interpretability and controllability of T2I generation.
\end{abstract}
\begin{keywords}
	Attention Maps, Attention Heads, Text-to-Image Generation
\end{keywords}

\section{Introduction}
Recently, text-to-image (T2I) generative models have achieved impressive results, producing high-quality images that closely align with input text prompts. Built on powerful foundation models such as Stable Diffusion~\cite{rombach2022high, podell2024sdxl} and Imagen~\cite{saharia2022photorealistic}, a wide range of works have extended these capabilities to tasks like image editing~\cite{hertz2022prompt, tumanyan2023plug, cao2023masactrl}, multi-concept generation~\cite{chefer2023attend, feng2022training}, and personalization~\cite{ruiz2023dreambooth, kumari2023multi}. Many of these approaches leverage architectural components—particularly cross-attention layers—to enhance performance and adaptability. However, despite this progress, studies specifically examining the roles and characteristics of different attention heads have remained largely limited.

A previous pioneering study~\cite{park2024cross} demonstrated that some cross-attention heads are more strongly associated with specific visual concepts than others, and that leveraging this information can improve tasks such as image editing and multi-concept generation. However, its focus was primarily limited to practical applications, leaving a gap between the model’s strong performance and a deeper understanding of the individual roles of attention heads. In this study, we aim to bridge that gap by comparing diffusion-based object segmentation results using only a selected subset of concept-relevant attention heads versus using all heads, as done in the widely adopted interpretability method DAAM~\cite{tang2022daam}. 

DAAM averages the cross-attention maps corresponding to a specific token and shows that the resulting map closely overlaps with the object referred to by that token. This demonstrates that averaged attention maps tend to focus on relevant objects, providing an intuitive explanation of how input tokens guide image generation. Building on this insight, attention map visualization has become a widely adopted tool for analyzing how individual tokens influence different regions of generated images. In this study, we extend this diffusion-based segmentation approach by selectively using only a few attention heads that are more strongly associated with the specified concept, aiming to enhance both interpretability and performance.

To identify attention heads related to specific concepts, we first apply the method proposed in HRV~\cite{park2024cross} to score each head based on its relevance to the given concept. We then select the top 20-25\% of attention heads with the highest relevance scores. The attention maps from this selected subset are averaged, and we compute the mean Intersection over Union (IoU) by comparing the result with ground-truth segmentation maps generated using Grounded-SAM~\cite{ren2024grounded}. Our approach outperforms the widely used DAAM method, suggesting that selectively chosen attention heads provide more accurate and concept-specific features.

As part of our analysis, we compared averaged cross-attention maps aggregated from a subset of attention heads most relevant to the concept `Animals’ with those from the least relevant heads. The significant performance gap between these two groups reveals that the most relevant attention heads more accurately capture concept-specific features than the least relevant ones. Additionally, we observe that when a T2I model misinterprets an input prompt—such as when ambiguous terms result in multiple concepts being generated in a single image—attention heads associated with different concepts often focus on distinct regions of the generated image, offering insight into the source of such errors. These findings highlight the potential of selective attention aggregation as a valuable tool for understanding and diagnosing the behavior of text-to-image generative models.

A large body of research in visual generative applications relies on attention maps as a core feature, and we believe that our selectively aggregated attention maps can enhance performance, improve visual interpretability, and enable finer control across these tasks.

\section{Method}
\subsection{Preliminaries}
\subsubsection{Cross-Attention Map of T2I Models. }
Let $\mathrm{P}$ be an input text prompt, $S$ the number of tokens in $\mathrm{P}$, and $\mathrm{Z}_t$ a noisy image latent at generation timestep $t$. In the $h$-th cross-attention head, the spatial features of the noisy image latent $\phi^{(h)} (\mathrm{Z}_t)$ are projected to a query matrix $\mathrm{Q}_t^{(h)} = l_Q^{(h)} (\phi^{(h)} (\mathrm{Z}_t)) \in \mathbb{R}^{r_h^2 \times d_k}$, while the CLIP text embedding of the input prompt $\psi (\mathrm{P})$ is projected to a key matrix $\mathrm{K}^{(h)} = l_K^{(h)} (\psi (\mathrm{P})) \in \mathbb{R}^{S \times d_k}$ and a value matrix $\mathrm{V}^{(h)} = l_V^{(h)} (\psi (\mathrm{P})) \in \mathbb{R}^{S \times d_v}$, using learned linear layers $l_Q^{(h)}, l_K^{(h)},$ and $l_V^{(h)}$. Here, $r_h^2$ denotes the spatial dimension (i.e., height $\times$ weight) for head $h$. Then, the cross-attention map at head $h$ and timestep $t$ is calculated as 
\begin{equation}
\label{eqn:cross-attention map calculation}
\mathrm{M}_t^{(h)} = \text{softmax} \left( \frac{\mathrm{Q}_t^{(h)} (\mathrm{K}^{(h)})^T}{\sqrt{d_k}} \right) \in \mathbb{R}^{r_h^2 \times S},
\end{equation}
where $d_k$ is the hidden dimension of the keys and queries. Each row of $\mathrm{M}_t^{(h)}$ corresponds to a spatial location and defines a probability distribution over the $S$ text tokens. Intuitively, $\mathrm{M}_t^{(h)}$ reveals which parts of the text are most relevant to each spatial location in the image, thereby guiding how semantic information from the prompt influences the image generation process at timestep $t$.

\subsubsection{DAAM~\cite{tang2022daam}. }
Motivated by the interpretability of the cross-attention map $\mathrm{M}_t^{(h)}$, the diffusion-based interpretability method DAAM proposes averaging attention maps across all attention heads and generation timesteps, resulting in the aggregated map: 
\begin{equation}
\label{eqn:daam}
\widehat{\mathrm{M}}_{DAAM} = \frac{1}{H} \sum_{h=1}^H F_h (\frac{1}{T} \sum_{t=1}^T  \mathrm{M}_t^{(h)}) \in \mathbb{R}^{r \times r \times S},
\end{equation}
where $H$ is the total number of attention heads in the T2I model, $T$ is the total number of generation timesteps, and $F_h(\cdot)$ denotes bicubic interpolation used to upscale each head’s spatial resolution $r_h^2$ to match the original image resolution $r \times r$. DAAM shows that this averaged attention map offers strong visual interpretability—it can be directly visualized for each token and provides meaningful insight into how different text tokens influence various spatial regions of the generated image. This underscores the usefulness of attention map analysis as a powerful technique for understanding and interpreting the behavior of T2I models.

\subsubsection{HRV~\cite{park2024cross}. }
HRV reveals that different visual concepts are processed unequally across different cross-attention heads, and proposes a method to quantify the relevance of each head to a set of human-specified visual concepts. Given a selected set of $N$ visual concepts $\{ C_1, \dots, C_N \}$, HRV first uses a state-of-the-art language model to generate 10 representative words--refer to as concept-words--for each concept, which are independently encoded using the CLIP text encoder $\psi(\cdot)$. During random image generation, one concept-word per visual concept is sampled, encoded, and projected into the key embedding space using the head-specific linear layer $l_K^{(h)}(\cdot)$, producing a concatenated matrix $\widetilde{\mathrm{K}}^{(h)} \in \mathbb{R}^{N \times d_k}$ of key-projected semantic token embeddings for the $N$ sampled concept-words at each cross-attention head position $h=1,\cdots,H$. Let $\mathrm{Q}_t^{(h)} \in \mathbb{R}^{r_h^2 \times d_k}$ be the spatial query features at timestep $t$ for head $h$ during the random image generation. Then, the attention map for HRV at head $h$ is computed as:
\begin{equation}
\label{eqn:HRV-relevance_score}
\widetilde{\mathrm{M}}_t^{(h)} = \text{softmax} \left( \frac{\mathrm{Q}_t^{(h)} (\widetilde{\mathrm{K}}^{(h)})^T}{\sqrt{d_k}} \right) \in \mathbb{R}^{r_h^2 \times N},
\end{equation}
where each column indicates how strongly each visual concept is activated at head $h$. These attention maps are averaged across spatial dimensions $r_h^2$ (i.e., height $\times$ width) to produce relevance score vector $s^{(h)} \in \mathbb{R}^N$. The most relevant concept at head $h$ is identified via  $n = \text{argmax} \left(s^{(h)}\right)$, and the $h$-th component of the head relevance vector $\bf{v}_n \in \mathbb{R}^H$, corresponding to concept $C_n$, is incremented by 1. After aggregating across all timesteps and heads over a sufficient number of random image generations, each concept's head relevance vector $\bf{v}_n$ is normalized to have unit $L_1$ norm. These head relevance vectors (HRVs) indicate which attention heads are most responsive to each visual concept and can be leveraged for interpreting, analyzing, or guiding generation in T2I models.

\subsection{Selective Attention Map Aggregation}
Based on the head relevance vector $\bf{v}_n$ obtained through HRV, we selectively choose a subset of attention heads that are most relevant to the concept of interest, depending on the goal of the analysis. We then propose to compute the averaged attention maps using only these selected heads, rather than aggregating across all heads as done in DAAM. Let $H_n$ denote the set of selected attention heads; the DAAM equation is then modified as follows:
\begin{equation}
\label{eqn:ours}
\widehat{\mathrm{M}}_{Ours} = \frac{1}{|H_n|} \sum_{h \in H_n} F_h (\frac{1}{T} \sum_{t=1}^T  \mathrm{M}_t^{(h)}) \in \mathbb{R}^{r \times r \times S}.
\end{equation}
In the following sections, we demonstrate the effectiveness of this modified analysis method and present several insightful findings derived using this approach.

\begin{table}[t]
\centering
\begin{tabular}{lrr}
\toprule
           & \multicolumn{1}{l}{Mean IoU score}  \\ \midrule
DAAM - 0.3 & 0.7490                                 \\
DAAM - 0.4 & 0.7540                                 \\
DAAM - 0.5 & 0.6261                                \\
Ours - 0.3 & 0.7698                                \\
Ours - 0.4 & \textbf{0.7765}                       \\
Ours - 0.5 & 0.6785                                \\ \bottomrule
\end{tabular}
\vspace{2mm}
\caption{Quantitative comparison of diffusion-based object semantic segmentation. The values following ``DAAM -'' or ``Ours -'' (i.e., 0.3, 0.4, 0.5) indicate the thresholds applied to the aggregated attention maps when computing the mean Intersection over Union (IoU).}
\label{tab:main_comparison}
\end{table}

\begin{figure}
    \centering
    \includegraphics[width=0.95\columnwidth,]{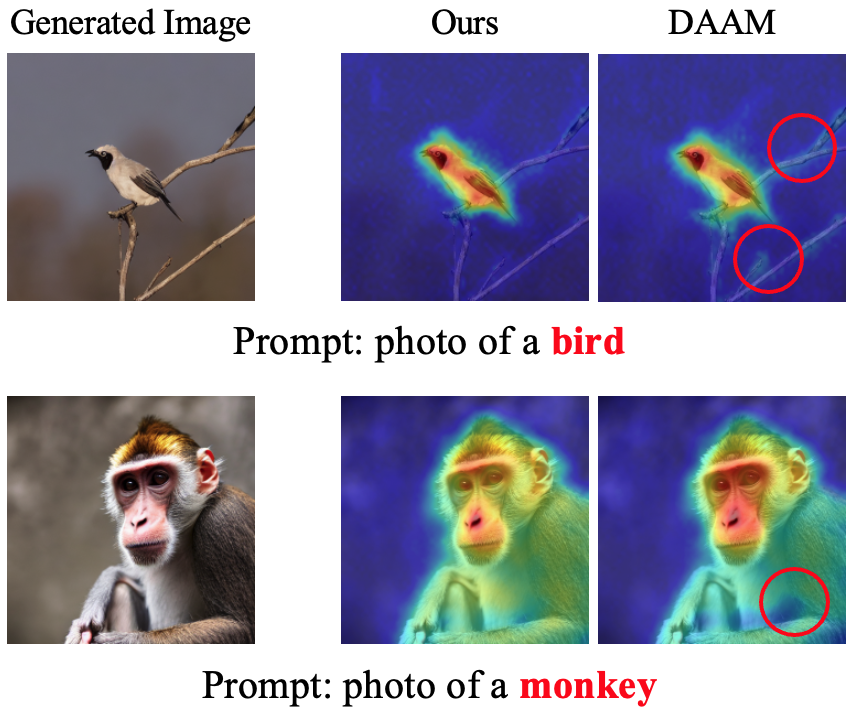}
    \caption{Qualitative comparison of diffusion-based object semantic segmentation. The original generated images are shown on the left, and the same images with aggregated attention maps overlaid are shown on the right. Red circles highlight regions with undesired focus or lack of focus.}
    \label{fig:ours_vs_daam}
\end{figure}

\section{Experiment}
\subsection{Experimental Setting}
We use Stable Diffusion v1.4 as the base model for both DAAM and our proposed method. Head relevance vectors are computed using a list of 34 visual concepts, each associated with 10 concept-words, as provided by the HRV code repository. For the benchmark focused on the Animals category, we adopt a simple prompt template: ``photo of a \{animal\}'', where the placeholder is replaced with one of the following 10 animal names: bear, bird, cat, cow, dog, elephant, sheep, horse, monkey, and zebra. For each prompt, we generate 10 images using different random seeds, resulting in a total of 100 generated images.

\begin{table}[t]
\centering
\begin{tabular}{lrr}
\toprule
  & \makecell[r]{Most Relevant \\ 30 Heads} & \makecell[r]{Least Relevant \\ 30 Heads} \\ \midrule
Ours - 0.3 & 0.7698                      & 0.6654                          \\
Ours - 0.4 & 0.7765                      & 0.6172                 \\
Ours - 0.5 & 0.6785                      & 0.4649                          \\ \bottomrule
\end{tabular}
\vspace{2mm}
\caption{Comparison of attention map aggregation using the most and least relevant attention heads. Aggregated attention maps are computed using either the 30 most or 30 least relevant attention heads for the concept `Animals.’ The values following ``DAAM -'' or ``Ours -'' (i.e., 0.3, 0.4, 0.5) indicate the thresholds applied to the aggregated attention maps when computing the mean Intersection over Union (IoU).}
\label{tab:most or least relevant heads}
\end{table}

\begin{figure}
    \centering
    \includegraphics[width=0.95\columnwidth,]{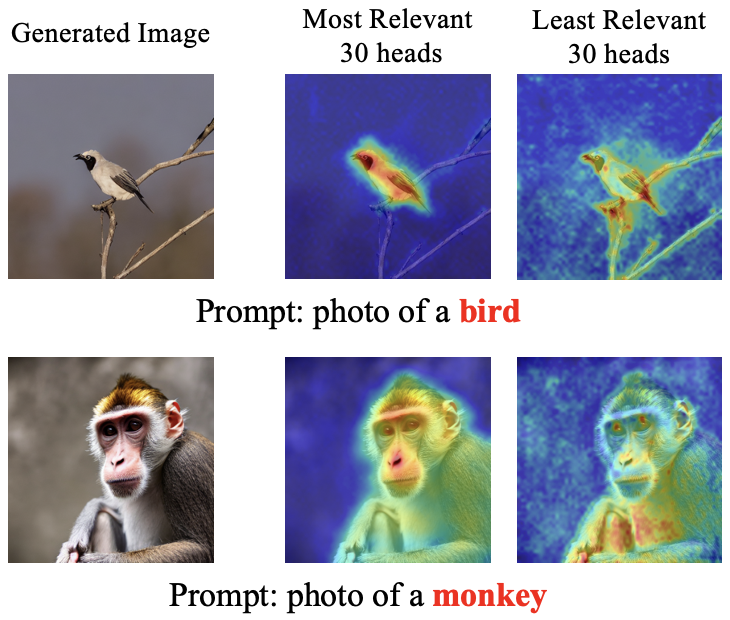}
    \caption{Visual comparison of attention map aggregation using the most and least relevant attention heads. The attention maps are computed using either the 30 most or 30 least relevant heads for the concept ‘Animals’ and are overlaid on the generated images to illustrate differences in focus regions.}
    \label{fig:related_vs_unrelated}
\end{figure}

\begin{figure*}
    \centering
    \includegraphics[width=0.8\textwidth,]{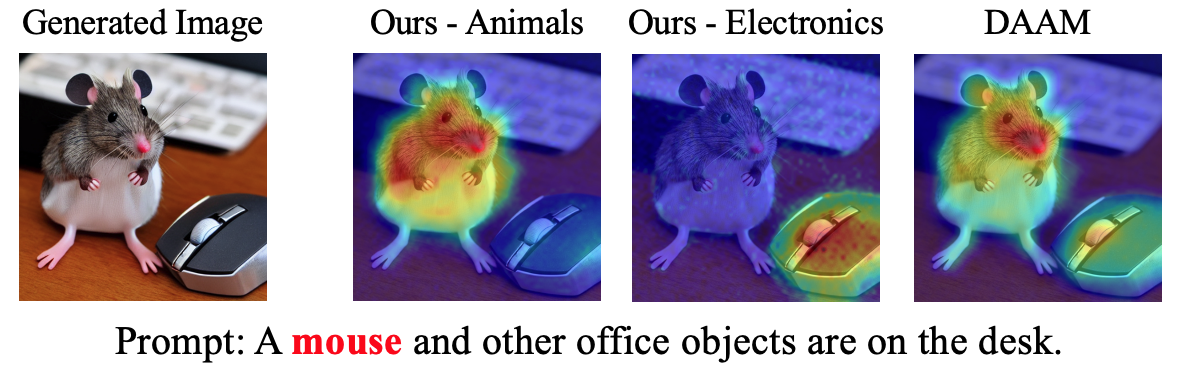}
    \caption{Visual diagnosis of text-to-image misinterpretation with selectively aggregated attention maps. The T2I model mistakenly generates both meanings of the word `mouse’--as an animal and as an electronic device. ‘Ours-Animals’ shows the aggregated attention map from the 30 heads most relevant to the concept `Animals,’ while ‘Ours-Electronics’ shows the map from the 30 heads most relevant to the concept `Electronics’. Note that DAAM aggregates attention maps from all heads in the T2I model. All three visualizations show the aggregated attention maps of the word `mouse.'}
    \label{fig:divergence_between_heads}
\end{figure*}

\subsection{Diffusion-Based Object Semantic Segmentation}
To evaluate the effectiveness of our selective attention map aggregation method, we compare it against DAAM using the benchmark focused on the Animals category. We generate 100 images using the standard Stable Diffusion v1.4 model and obtain corresponding ground-truth semantic segmentation maps with Grounded-SAM. For each image, we aggregated attention maps which are captured during generation process—one using DAAM, as defined in Eq.~\ref{eqn:daam}, and the other using our method, which averages the 30 attention heads that are most relevant to the concept `Animals,’ as defined in Eq.~\ref{eqn:ours}. To quantitatively compare these maps with the ground-truth segmentations, we apply thresholding at values of 0.3, 0.4, or 0.5 to convert the continuous attention maps into binary masks. We then compute the Intersection over Union (IoU) score as
\begin{equation}
\text{IoU}_{v} (\widehat{\mathrm{M}}[:,;,s], \mathrm{G}) = \frac{\text{Area}(e_{v} (\widehat{\mathrm{M}}[:,:,s]) \cap \mathrm{G})}{\text{Area}(e_{v}(\widehat{\mathrm{M}}[:,:,s]) \cup \mathrm{G})},
\end{equation}
where $e_{v}$ is the thresholding operation with threshold $v \in \{0.3, 0.4, 0.5\}$, $\widehat{\mathrm{M}}$ represents either $\widehat{\mathrm{M}}_{DAAM}$ or $\widehat{\mathrm{M}}_{Ours}$, $s$ denotes the token corresponding to the animal name, and $\mathrm{G}$ is the ground-truth segmentation map. The final performance is reported as the mean IoU averaged over all 100 generated images.

Table~\ref{tab:main_comparison} presents the quantitative results of DAAM and our method across three different threshold values. At all thresholds, our approach consistently achieves higher mean IoU scores, indicating that the aggregated attention maps produced by our method more closely align with the ground-truth segmentation maps. Figure~\ref{fig:ours_vs_daam} provides qualitative comparisons by visualizing the aggregated attention maps overlaid on the generated images. The red circles in the DAAM results highlight regions of undesired focus or lack of focus, illustrating that our method more accurately captures the target objects within the generated images.

\section{Analysis}
\subsection{Most and Least Relevant Attention Heads}
We compare aggregated attention maps derived from the most relevant 30 attention heads to those from the least relevant 30 heads to effectively demonstrate the importance of selective aggregation. As shown in Table~\ref{tab:most or least relevant heads}, using the least relevant heads for the concept `Animals’ results in a significant drop in mean IoU compared to using the most relevant heads. Figure~\ref{fig:related_vs_unrelated} further demonstrates that these least relevant heads tend to focus on regions outside of the animals, providing evidence that applying adaptation techniques indiscriminately across all attention maps or heads may be suboptimal for various applications.

\subsection{Visual Diagnosis of Text-to-Image Misinterpretations}
T2I models often misinterpret words in input prompts, either due to inherent limitations of the model or ambiguities in the prompts themselves. We find that some of these misinterpretations can be visually explained using our selectively aggregated attention maps. In Figure~\ref{fig:divergence_between_heads}, given the input prompt `A mouse and other office objects are on the desk,’ the T2I model interprets the word `mouse’ as both an animal and an electronic device, generating both concepts in a single image. When we visualize the aggregated attention maps using the 30 attention heads most relevant to the concept ‘Animals,’ the focus is solely on the mouse as an animal. In contrast, using the 30 heads most relevant to ‘Electronics’ highlights only the electronic mouse. DAAM, which uses all attention heads, shows focus on both objects. This example demonstrates that selective attention map aggregation can offer a new way to visually interpret and diagnose misinterpretations in text-to-image generations.

\subsection{Ablation Study on the Number of Selected Heads}
Our base model, Stable Diffusion v1, contains 128 cross-attention heads, and our method includes a hyperparameter controlling how many heads to select. To determine an appropriate number, Table~\ref{tab:ablating head numbers} presents an ablation study on this parameter using the Animals category benchmark. The results show that selecting 30 heads achieves the best performance, so we consistently use this number throughout the paper.

\begin{table}[t]
\centering
\begin{tabular}{lrrr}
\toprule
 & \multicolumn{1}{l}{20 heads} & \multicolumn{1}{l}{30 heads} & \multicolumn{1}{l}{40 heads} \\ \midrule
Ours - 0.3 & 0.7436            & 0.7698          & 0.7669                \\
Ours - 0.4 & 0.7001            & \textbf{0.7765}          & 0.7412                \\
Ours - 0.5 & 0.5386            & 0.6785          & 0.6315                \\ \bottomrule
\end{tabular}
\vspace{2mm}
\caption{Ablation study on the number of selected attention heads. The values following ``DAAM -'' or ``Ours -'' (i.e., 0.3, 0.4, 0.5) indicate the thresholds applied to the aggregated attention maps when computing the mean Intersection over Union (IoU). Note that Stable Diffusion v1 contains 128 cross-attention heads. }
\label{tab:ablating head numbers}
\end{table}

\section{Limitation and Future Work}
Our study has several limitations regarding the scope of the model and data selection. Although we evaluated our approach only on Stable Diffusion v1.4, we expect that our findings can be generalized to other text-to-image models, as they similarly rely on attention heads as key components--an expectation supported by prior research on the roles of different attention heads across various T2I models~\cite{park2024cross, ahn2025fine}. For experimental convenience, we focused exclusively on the concept of `Animals,' but expanding our approach to encompass a broader range of visual concepts is a critical next step. Building on these preliminary results, future work will aim to apply our method to diverse T2I architectures and broaden its applicability to various visual concepts.

\section{Conclusion}
In this work, we present preliminary findings on selectively aggregating attention maps to enhance diffusion-based visual interpretations. Our approach outperforms the widely used DAAM technique in diffusion-based segmentation and visually demonstrates that the most relevant attention heads effectively capture the corresponding concepts within the attention maps, while the least relevant heads do not. Additionally, we provide a case study showing how selective attention aggregation can visually diagnose misinterpretations in text-to-image models. Overall, our results suggest that selectively leveraging attention maps, rather than aggregating them indiscriminately, leads to enhanced visual interpretability and broader applicability.

\section*{Acknowledgements}
This work was partly supported by Institute of Information \& communications Technology Planning \& Evaluation (IITP) grant funded by the Korea government (MSIT) ([NO.RS-2021-II211343, Artificial Intelligence Graduate School Program (Seoul National University)], [No. RS-2023-00235293, Development of autonomous driving big data processing, management, search, and sharing interface technology to provide autonomous driving data according to the purpose of usage]).

\bibliography{references}

\end{document}